\documentclass[conference]{IEEEtran}
\IEEEoverridecommandlockouts
\usepackage{cite}
\usepackage{amsmath,amssymb,amsfonts}
\usepackage{algorithm} 
\usepackage{graphicx}
\usepackage{caption}
\usepackage{subcaption}
\usepackage{textcomp}
\usepackage{xcolor}
\usepackage{array}
\usepackage{textcomp}
\usepackage{stfloats}
\usepackage{url}
\usepackage{verbatim}
\usepackage{amsthm, bm, hyperref, mathrsfs, indentfirst, longtable}
\usepackage{svg}
\usepackage{longtable}
\usepackage{tabularx}
\usepackage{makecell}
\usepackage{booktabs,float}
\usepackage{bbm}
\usepackage{balance}
\usepackage{threeparttable}
\usepackage{multirow}
\usepackage{booktabs}
\def\BibTeX{{\rm B\kern-.05em{\sc i\kern-.025em b}\kern-.08em
    T\kern-.1667em\lower.7ex\hbox{E}\kern-.125emX}}
\begin{document}
\title{BeamLLM: Vision-Empowered mmWave Beam Prediction with Large Language Models\\
}

\author{Can Zheng$^{1}$, Jiguang~He$^2$, Guofa Cai$^3$, Zitong Yu$^2$, Chung G. Kang$^1$\\
        $^1$School of Electrical Engineering, Korea University, Seoul, Republic of Korea\\
        $^2$School of Computing and Information Technology, Great Bay University, Dongguan 523000, China\\ 
        $^3$School of Information Engineering, Guangdong University of Technology, Guangzhou, China\\
        }
 \maketitle 

\maketitle
\begin{abstract}
In this paper, we propose BeamLLM, a vision-empowered millimeter-wave (mmWave) beam prediction framework leveraging large language models (LLMs) to enhance the accuracy and robustness of beam prediction. By integrating computer vision (CV) with LLMs’ cross-modal reasoning capabilities, the framework extracts user equipment (UE) positional features from RGB images and aligns visual-temporal features with LLMs’ semantic space through reprogramming techniques. Evaluated on a realistic vehicle-to-infrastructure (V$2$I) scenario, BeamLLM achieves $61.01\%$ top-$1$ accuracy and $97.39\%$ top-$3$ accuracy in standard prediction tasks, outperforming traditional deep learning models. In few-shot prediction scenarios, performance degradation is limited to $12.56\%$ (top-$1$) and $5.55\%$ (top-$3$) from time sample $1$ to $10$, demonstrating superior prediction capability.
\end{abstract}


\section{Introduction}
 
    Millimeter-wave (mmWave) communication offers abundant spectrum above $26$ GHz, enabling high data rates but suffering from severe path loss. To address this, large antenna arrays employ narrow beams to enhance link quality. Effective beamforming requires accurate beam alignment between transmitters and receivers, traditionally achieved via beam training using predefined codebooks. However, this process incurs high overhead, especially in high-mobility scenarios such as V$2$X and UAVs. To address this issue, a promising approach is to proactively predict beams by leveraging user-side positioning data and base station (BS) sensing measurements. For instance, RGB images captured by BS cameras provide both spatial location of the UE and rich environmental cues. Predicting beams from historical image sequences involves capturing the complex spatiotemporal interactions within sensing data-making. This task well-suited for DL, which can uncover hidden dynamics and learn mappings between visual input and future beam directions \cite{RGB, Radar, LiDAR, GPS}. As a key technique for integrated sensing and communication (ISAC) in $6$G, this approach shows strong potential for efficient mmWave MIMO operation.
    
    Recent breakthroughs in large language models (LLMs), such as GPT-$4$ \cite{GPT-4} and DeepSeek \cite{DeepSeek}, have demonstrated remarkable capabilities in contextual reasoning and few-shot generalization abilities. While LLMs are originally designed for natural language processing (NLP), LLMs have shown strong cross-modal learning capabilities, thus extending their applications to other tasks. Inspired by these advantages of the LLMs, several works applying LLMs have been proposed for channel prediction\cite{LLMCP}, beam prediction \cite{LLMBP}, and port prediction for fluid antennas\cite{LLMPP}.

    Built on these developments, in this paper, we propose a vision-empowered beam prediction framework, named BeamLLM, which utilizes LLMs to process historical RGB image series, thereby enabling more efficient and adaptive beam prediction. Unlike \cite{LLMBP}, our method does not rely on historical beam indices or angle of departure (AoD) information. Instead, BeamLLM relies solely on visual features for beam prediction. Furthermore, to ensure robust performance and practical applicability, we validate our framework using real-world measurement datasets, to demonstrate its potential for deployment in real-world scenarios.

    The rest of this paper is organized as follows: Section II provides a system model and problem formulation of the beam prediction task. The proposed BeamLLM framework is presented in Section III. Section IV presents simulation results, including performance comparisons with benchmark methods, along with detailed discussions. Finally, we conclude our work in Section V.

    \textit{Notations}: Bold lowercase letters denote vectors (e.g., $\mathbf{x}$), and bold uppercase letters denote matrices (e.g., $\mathbf{X}$). The superscripts $(\cdot)^\mathsf{T}$ and $(\cdot)^\mathsf{H}$ represent the transpose and Hermitian (conjugate transpose) operations, respectively. The operator $\mathbb{E}[\cdot]$ denotes the statistical expectation, while $|\cdot|_2$ denotes the Euclidean norm of a vector, and $|\cdot|$ returns the magnitude of a complex number. The indicator function $\mathbbm{1}\{\cdot\}$ equals $1$ if the condition inside the braces is true, and $0$ otherwise. Unless otherwise specified, $\mathbb{Z}^{+}$, $\mathbb{R}$, and $\mathbb{C}$ denote the sets of positive integer, real, and complex numbers, respectively.

    \begin{figure}[ht]
        \centering
        \includegraphics[width=\linewidth]{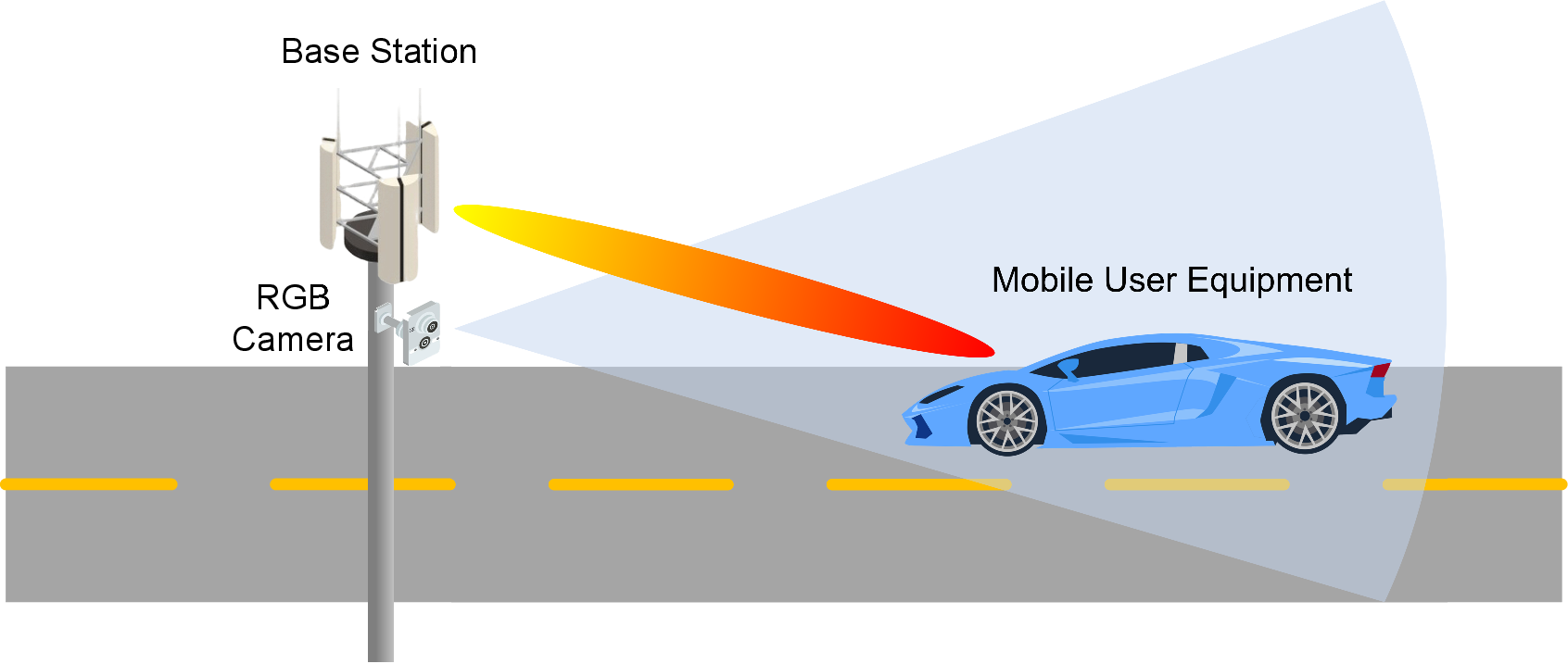}
        \caption{Illustration of the system model, showing a camera-equipped BS communicating with a moving UE, while RGB images are collected to support communications.}
        \label{fig:structure}
    \end{figure}  

\section{System Model and Problem Formulation}
    \subsection{System Model}

    Fig. \ref{fig:structure} illustrates the system model considered for vehicle-to-infrastructure (V$2$I) mmWave communication. In this model, the BS deploys a mmWave phased-array receiver with $N$ elements of half-wavelength spacing and an RGB camera. The antenna array enables the BS to perform beamforming, while the camera captures images within its field of view at a certain frame rate for sensing and downstream applications. 
    
    \begin{figure*}[ht]
        \centering
        \includegraphics[width=0.9\textwidth]{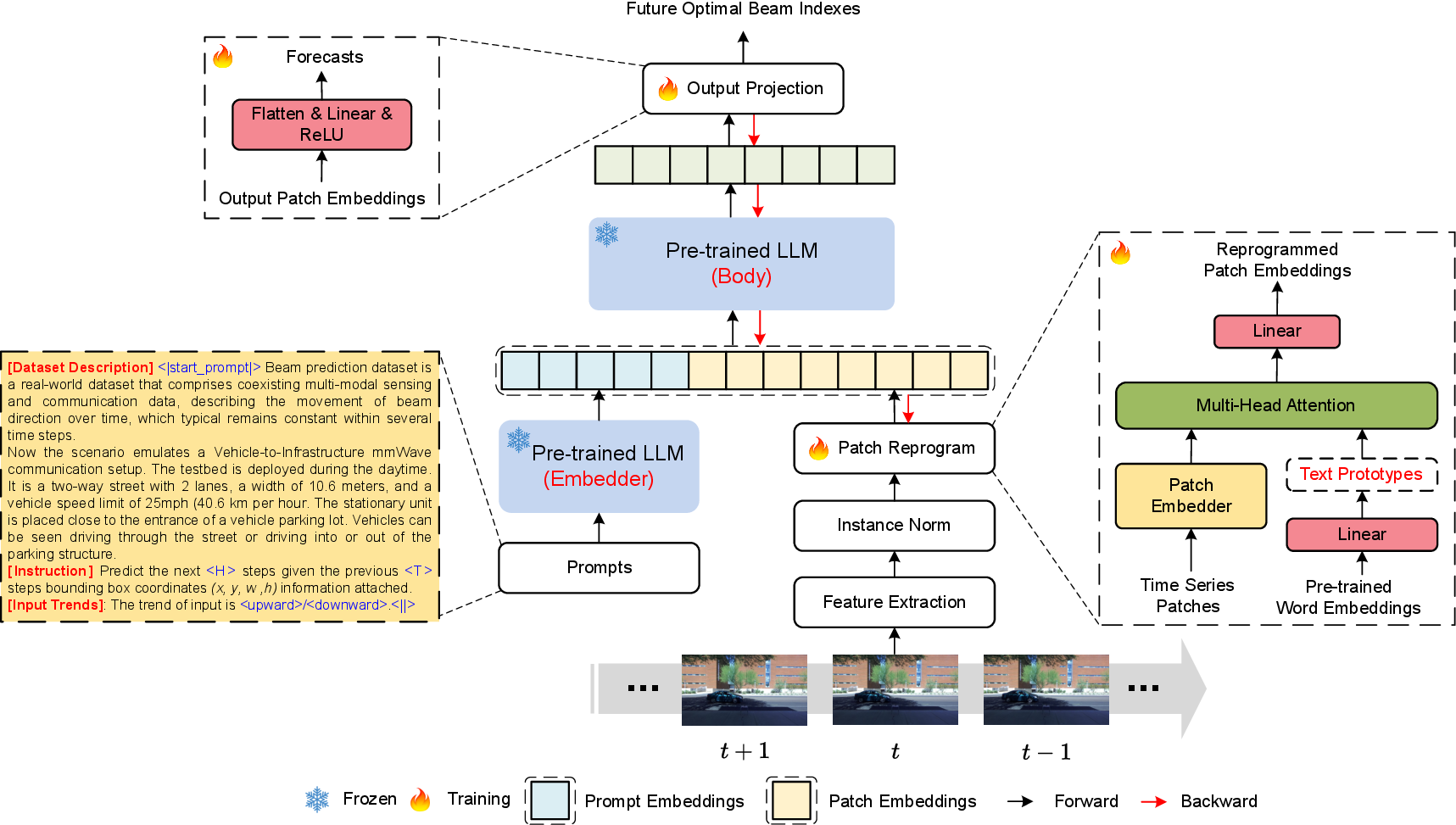}
        \caption{The model framework of BeamLLM.}
        \label{fig:BeamLLM}
    \end{figure*}
    
    We assume that the BS has a predefined beamforming codebook $\mathcal{F} = \{\mathbf{f}_1, \cdots, \mathbf{f}_M\}$, containing $M$ beams, where $\mathbf{f}_m\in \mathbb{C}^{N\times 1}, m = 1,\cdots, M$ represents the $m$-th beamforming vector. The UE is assumed to have a single antenna. At time step $t$, the user transmits a single symbol $s[t]\in \mathbb{C}$ that satisfies the power constraint $\mathbb{E}[|s[t]|^2] = P$, where $P$ represents the transmit power. At the BS, the received signal $y[t]$ can be expressed as:
    \begin{align}
        y[t] = \mathbf{h}^\mathsf{H}[t] \mathbf{f}_{m[t]}s[t] + n[t],
    \end{align}
    where $\mathbf{h}[t]$ is the channel vector, $\mathbf{f}_{m[t]}$ is the $m[t]$-th beamforming vector from the codebook in time step $t$, and $n[t]\sim \mathcal{CN}(0,\sigma^2)$ is the additive white Gaussian noise (AWGN) with variance $\sigma^2$.
    
    \subsection{Problem Formulation}
    This paper mainly focuses on the beam prediction problem at the BS. Given the available sensing information up to time $t-1$, the BS attempts to determine the optimal beams for $T_\text{pred}\in \mathbb{Z}^{+}$ future time steps, specifically for $t,\cdots,(t+T_\text{pred} -1)$. We define the optimal beam at time step $t$ as the beam that provides the highest beamforming gain, given by:
    \begin{align}
        \mathbf{f}_{m^{*}[t]} = \arg \max_{\mathbf{f}_{m[t]} \in \mathcal{F}} |\mathbf{h}^\mathsf{H}[t]\mathbf{f}_{m[t]}|^2.
    \end{align}

    When perfect CSI knowledge is unavailable, beam training serves as an alternative method for determining the optimal beam. However, with a narrow beam codebook, the training overhead can be significant, and the likelihood of identifying the optimal beam is often low when the pre-beamforming signal-to-noise ratio (SNR) is poor. Because the optimal beam selection at the transmitter and receiver depends on the surrounding environment of the transceiver, our work aims to leverage visual information at the BS to assist beam selection and develop a beam prediction framework. Cameras deliver detailed visual data without requiring wireless resources or feedback signals, unlike radar or wireless-based sensing. As mature, cost-effective sensors, they are easier to implement and deploy. Additionally, cameras can leverage advanced computer vision algorithms for enhanced functionality \cite{RGB}.

\section{Large Laguage Model-Based Beam Prediction}
    In this section, we introduce BeamLLM to tackle the vision-assisted beam prediction task outlined in Section II. Fig. \ref{fig:BeamLLM} illustrates the proposed BeamLLM. The architecture mainly comprises two components, i.e., the visual data feature extraction module and the backbone module. 

    \subsection{Visual Data Feature Extraction Module}

    To process raw RGB data for the vision-aided beam prediction task, we employ the YOLOv4 object detector\cite{YOLOv4}. We select YOLOv4 due to its proven stability, computational efficiency, and robust performance in real-time object detection tasks, which are critical for practical mmWave beam prediction scenarios. This detector identifies potential UEs within RGB images and extracts bounding box vectors $\mathbf{b}$. For a single image $\mathbf{X}_I\in \mathbb{R}^{W_\text{I} \times H_\text{I} \times C_\text{I}}$, where $W_I$, $H_I$ and $C_I$ represent the spatial width, height, and RGB/IR spectral channels respectively, the bounding box vector is given as:
    \begin{align}
        \mathbf{b}=\text{YOLO}(\mathbf{X}_I) = [x_c,y_c,w,h]^\mathsf{T},
    \end{align}
    which consists of the detected object's center coordinates ($x$-axis, $y$-axis), width, and height within the RGB image. Since the optimal beam selection is highly dependent on the direction and position of the transmission target, we use a sequence of bounding box vectors as the extracted visual feature. The objective is to predict the optimal beam index for the next $T_\text{pred}$ steps based on the historical $T_\text{hist}$ step bounding box vectors, denoted as $\mathbf{B} = [\mathbf{b}[t-T_\text{hist}], \cdots, \mathbf{b}[t-1]]\in \mathbb{R}^{4 \times T_\text{hist}}$.

    \subsection{The Backbone Module} 
    The inherent potential of LLMs can be utilized to address the beam prediction task. However, a key challenge lies in aligning the visual feature modality with the textual modality to enable LLMs to effectively comprehend the task. Furthermore, training LLMs requires extensive datasets, which is often unrealistic in practical scenarios.
    
    Time-LLM simultaneously addresses both challenges through reprogramming \cite{TimeLLM}, which consists of two key steps: \textbf{adaptation} and \textbf{alignment}. Specifically, adaptation is achieved via the patch reprogramming module, which enables LLMs to process input data effectively, thereby breaking domain isolation and facilitating knowledge sharing. Alignment, on the other hand, is accomplished through the \textit{prompt-as-prefix} (PaP) module, which further eliminates domain boundaries to enhance knowledge acquisition.

    \textbf{Input Embedding:} For each row of $\mathbf{B}$, denoted as $\mathbf{B}^{(i)}\in \mathbb{R}^{1\times T_\text{hist}}$ for $i=1,2,3,4$, reversible instance normalization (RevIN)\cite{RevIN} is applied individually to normalize the data, ensuring a mean of $0$ and a variance of $1$. RevIN dynamically adjusts the normalization parameters to accommodate variations in the data distribution. Subsequently, $\mathbf{B}^{(i)}$ is segmented into several contiguous overlapping or non-overlapping patches, each of length $L_p$. The total number of input patches is given by $\lfloor\frac{T_\text{hist}-L_p}{S}\rfloor+2$, where $S$ represents the horizontal sliding size. This operation is inspired by techniques in CV, wherein local temporal information is aggregated within each patch to better preserve local semantic features. Finally, a simple linear layer is employed to embed $\mathbf{B}^{(i)}_P\in \mathbb{R}^{P\times L_p}$ into $\hat{\mathbf{B}}^{(i)}_P\in \mathbb{R}^{P\times d_m}$.

    \textbf{Patch Reprogramming:} To enable LLMs to process visual features $\hat{\mathbf{B}}^{(i)}_P$, which differ from natural language in modality, a reprogramming layer maps temporal input features to an NLP task using cross-attention \cite{CAT}. A linear layer projects pre-trained word embeddings $\mathbf{E} \in \mathbb{R}^{V \times D}$ onto a smaller set of text prototypes $\mathbf{E}' \in \mathbb{R}^{V' \times D}$, where $V$ and $V'$ denote the vocabulary sizes of $\mathbf{E}$ and $\mathbf{E}'$, respectively $(V' \ll V)$,  and $D$ is the hidden dimension. For each attention head $k=1,\ldots,K$, we compute:
    
    \begin{align}
        \mathbf{Q}_{k}^{(i)} &= \hat{\mathbf{B}}^{(i)}_P \mathbf{W}_{k}^{Q}, \\
        \mathbf{K}_{k}^{(i)} &= \mathbf{E}' \mathbf{W}_{k}^{K}, \\
        \mathbf{V}_{k}^{(i)} &= \mathbf{E}' \mathbf{W}_{k}^{V},
    \end{align}
    where $\mathbf{W}^Q_k\in \mathbb{R}^{d_m\times \lfloor \frac{d_m}{K} \rfloor}$ and $\mathbf{W}^K_k, \mathbf{W}^V_k\in \mathbb{R}^{D\times \lfloor \frac{d_m}{K} \rfloor}$. The following process adaptively obtains the text descriptions corresponding to patches through a multi-head self-attention mechanism:
    \begin{align}
        \mathbf{Z}_{k}^{(i)} &= \mathrm{ATTENTION}\left( \mathbf{Q}_{k}^{(i)}, \mathbf{K}_{k}^{(i)}, \mathbf{V}_{k}^{(i)}\right)\nonumber \\
        &= \mathrm{SOFTMAX}\left(\frac{\mathbf{Q}_{k}^{(i)} {\mathbf{K}_{k}^{(i)}}^{\top}}{\sqrt{d_k}}\right) \mathbf{V}_{k}^{(i)},
    \end{align}
    where $\mathrm{ATTENTION}(\cdot)$ represents the scaled dot-product attention mechanism and $\mathrm{SOFTMAX}(\cdot)$ normalizes attention scores into a probability distribution, respectively. By aggregating each $\mathbf{Z}^{(i)}_k \in \mathbb{R}^{P \times d}$ across all heads, we obtain $\mathbf{Z}^{(i)} \in \mathbb{R}^{P \times d_m}$. This is then linearly projected to align the hidden dimension with the backbone model, resulting in $\mathbf{O}^{(i)} \in \mathbb{R}^{P \times D}$.

    \textbf{PaP:} Natural language-based prompts serve as prefixes to enrich the input context and guide the transformation of reprogrammed patches. We have identified three essential components for constructing an effective prompt: (1) dataset description, (2) task description, and (3) input statistics. The dataset description offers the LLM with fundamental background information about the input features, which often exhibit distinct characteristics across different domains. The task description offers crucial guidance to the LLM for transforming patch embeddings in the context of the specific task. Additionally, we incorporate supplementary key statistics, such as trends, to further enrich the input features, facilitating pattern recognition and reasoning.

    \textbf{Output Projection:} By packaging and forwarding the prompts along with the patch embeddings $\mathbf{O}^{(i)}$ through the frozen LLM, we discard the prefix portion and obtain the output representations. These representations are then flattened and linearly projected to produce the final outputs, $\hat{\mathbf{P}} = [\hat{\mathbf{p}}[t], \cdots, \hat{\mathbf{p}}[t+T_\text{pred}-1]]\in \mathbb{R}^{M\times T_\text{pred}}$. The index of the dimension corresponding to the maximum value of each $\hat{\mathbf{p}}[t]$, is predicted as the optimal future beam index, given by:
    \begin{align}
        \hat{m}^{*}[t] = \arg \max_{m}  \hat{\mathbf{p}}[t].
    \end{align}

    \subsection{Learning Phase}
    As shown in Fig. \ref{fig:BeamLLM}, certain modules are marked with a flame icon, indicating that they are trainable components. In contrast, the LLM backbone is marked with a snowflake icon, signifying that its parameters are frozen during training. The model is trained in a supervised manner using image inputs paired with the corresponding ground truth optimal beam indices. The beam prediction task is essentially a time-series forecasting task, where each time step corresponds to a classification task; therefore, the model parameters are optimized by minimizing the cross-entropy, which is expressed as:
    \begin{align}
        \mathcal{L}=-\sum_{j=t}^{t+T_\text{pred}-1}\sum_{m=1}^{M} \mathbbm{1}\{\hat{m}^*[j] = m^*[j]\}\log_2 (\hat{p}_m[j]),
    \end{align}
    where 
    $\hat{p}_m[j]$ is the $m$-th element of the output vector $\hat{\mathbf{p}}[j]$ at time step $j$, respectively.

\section{Performance Evaluation}

    We utilize the DeepSense $6$G dataset\cite{deepsense6g} for simulation and performance evaluation. DeepSense $6$G is a multimodal dataset from real-world measurements, including wireless beam data, RGB images, GPS locations, radar, and LiDAR.

    \subsection{Experimental Settings}

    \textbf{Dataset Processing:} We adopt Scenario $8$ of the DeepSense $6$G dataset for our simulation, which simulates a V$2$I mmWave communication setup. The BS is equipped with an RGB camera and a $16$-element $60$ GHz mmWave phased array, while the mobile UE serves as a mmWave transmitter. During data collection, the UE passes by the BS multiple times. At each time step, the BS captures an RGB image of the UE while scanning all predefined beams and measuring the received power for all $M = 32$ beams in a codebook. The multimodal data streams are synchronized to ensure temporal consistency.

    The dataset is split into $70\%$ training, $10\%$ validation, and $20\%$ test sets. The dataset consists of multiple data sequences. Each data sequence is a pair comprising an RGB image sequence and a beam index sequence. For each data sequence, we decompose it into data samples using a sliding window of size $13$.  As previously mentioned, during training, we use an observation window of size $T_\text{hist}$, and we train the model to predict future beams over a horizon $T_\text{pred}$. Therefore, the input to the encoder for the model is $\mathbf{X}_I[t-T_\text{hist}], \ldots, \mathbf{X}_I[t-1]$. In both beam prediction methods, the expected output from the decoder is $\hat{\mathbf{p}}[t], \ldots, \hat{\mathbf{p}}[t+T_\text{pred}-1]$. Since we maintain a fixed sequence length of $13$, we set $T_\text{hist} = 8$, $T_\text{pred} = 5$ as standard prediction and $T_\text{hist} = 3$, $T_\text{pred} = 10$ as few-shot prediction.

    \textbf{Baselines:} We compare our approach with several classical time-series models, including RNN (the implementation is consistent with \cite{RGB}), GRU, and LSTM. Additionally, to validate the effectiveness of the PaP module, we conduct an ablation study by comparing our model with and without PaP in the standard prediction setup.
    
    \textbf{Parameter Settings:} BeamLLM is configured as following: 1) A widely-used language model, i.e., GPT-$2$\cite{GPT-2}, is employed as the LLM backbone; 2) It is trained with Adam optimizer, where the batch size and initial learning rate (LR) are $16$ and $0.01$, respectively. Additionally, a multi-step LR scheduler in $1$, $5$, $10$, $15$, $20$, $25$, $30$, $40$ epochs with a decay factor of $\gamma = 0.9$ is employed; 3) The training process is set to $200$ epochs. The detailed model parameters are shown in Table 
    \ref{tab:comparison}.
    
    \begin{table}[ht]
    \caption{Parameter settings of different models.}
    \centering
    \begin{tabular}{|l|l|l|}
    \hline
    \textbf{Component} & \textbf{BeamLLM} & \textbf{RNN/GRU/LSTM} \\ \hline
    Input Layers & \begin{tabular}[c]{@{}l@{}}Patch Reprogramming\\ (Same as \cite{TimeLLM}, $V'=64$)\end{tabular} & Linear $4\times32$ \\ \hline
    Output Layers & \begin{tabular}[c]{@{}l@{}}Output Projection:\\ 1. Linear $4\times16$ + ReLU\\ 2. Linear $16\times32$ + ReLU\\ 3. Linear $32\times32$\end{tabular} & \begin{tabular}[c]{@{}l@{}}Sequence Model:\\ Layers $1$-$4$: $32\times32$\end{tabular} \\ \hline
    \end{tabular}
    \label{tab:comparison}
    \end{table}

    \textbf{Performance Metrics:} Top-$K$ accuracy is a metric that quantifies the percentage of validation samples for which the best ground truth beam is among the top $K$ model predictions with the highest probability. Mathematically, it is represented as:

    \begin{align}
        \text{Top-}K\ \text{accuracy} = \frac{1}{N_\text{test}}\sum_{i=1}^{N_\text{test}} \mathbbm{1}{\{m_i\in Q_k\}},
    \end{align}
    where $N_\text{test}$ represents the total number of samples in the test set, $m_i$ denotes the index of the ground truth optimal beam for the $i$-th sample, and $Q_k$ is the set of indices for the top-$K$ predicted beams, sorted by the element values in $\hat{\mathbf{P}}$ for each time sample.

    \subsection{Dataset Visualization}
    
    \begin{figure}[t]
        \centering
        \includegraphics[width=\linewidth]{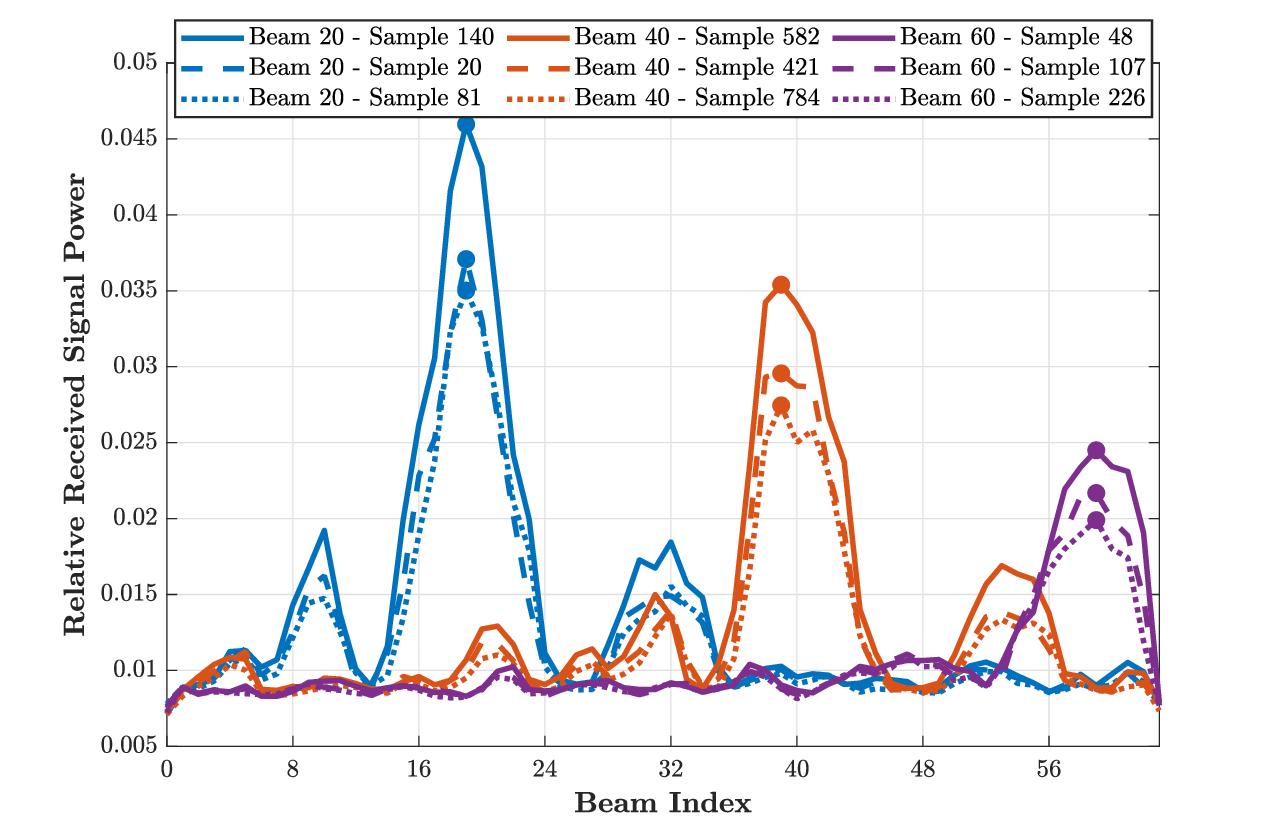}
        \caption{Comparison of power profiles for the first $3$ samples associated with optimal beams $20$, $40$, and $60$, respectively.}
        \label{fig:power}
    \end{figure}

    \begin{figure}[!t]
        \centering
          \begin{subfigure}{\linewidth}
          \centering
            \includegraphics[width=\textwidth]{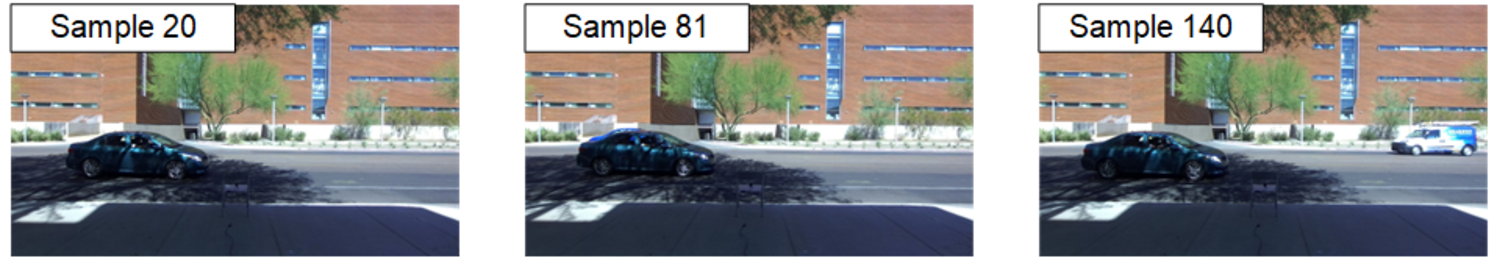}
            \caption{Vision data visualization for the first $3$ samples associated with optimal beam $20$.}
            \label{fig:20}
          \end{subfigure}
          \vspace{0.5em}
          \begin{subfigure}{\linewidth}
          \centering
            \includegraphics[width=\textwidth]{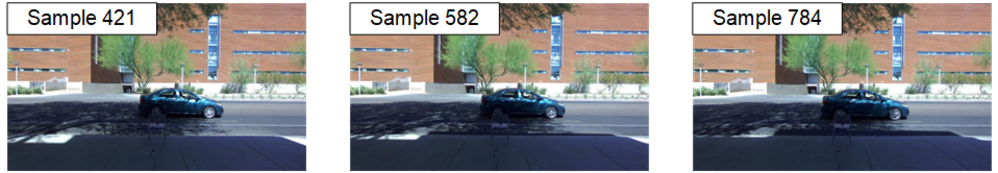}
            \caption{Vision data visualization for the first $3$ samples associated with optimal beam $40$.}
            \label{fig:40}
          \end{subfigure}
          \vspace{0.5em}
          \begin{subfigure}{\linewidth}
          \centering
            \includegraphics[width=\textwidth]{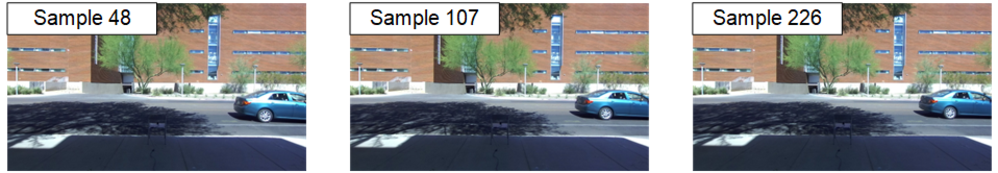}
            \caption{Vision data visualization for the first $3$ samples associated with optimal beam $60$.}
            \label{fig:60}
          \end{subfigure}
          \caption{Vision data visualization for the first $3$ samples associated with optimal beams $20$, $40$, and $60$, respectively.}
          \label{fig:image}
    \end{figure}

    Fig. \ref{fig:power} illustrates the relative received power distribution across various beam indices ($20$, $40$, $60$) for the first three samples, highlighting how peak power varies with different beam selections. Note that due to the complexity of absolute power calibration, the dataset provides dimensionless relative received signal power \cite{deepsense6g}. From the figure, it can be seen that as the optimal beam index increases, the relative received power shows an overall increasing trend. This implies that the distance between transceivers in the dataset may be smaller when the beam index is larger, leading to an increase in received power. Fig. \ref{fig:image} shows the first three image samples when the optimal beam is $20$, $40$, and $60$, respectively. We can see that as the blue car travels from the left side of the screen to the right, the optimal beam index progressively increases. This suggests that the optimal beam is highly dependent on the direction and position of the UE, and that visual information can intuitively provide information on the relative orientation position of the UE, as well as additional environmental information.

    \subsection{Convergence Analysis}

\begin{figure}[t]
        \centering
        \includegraphics[width=\linewidth]{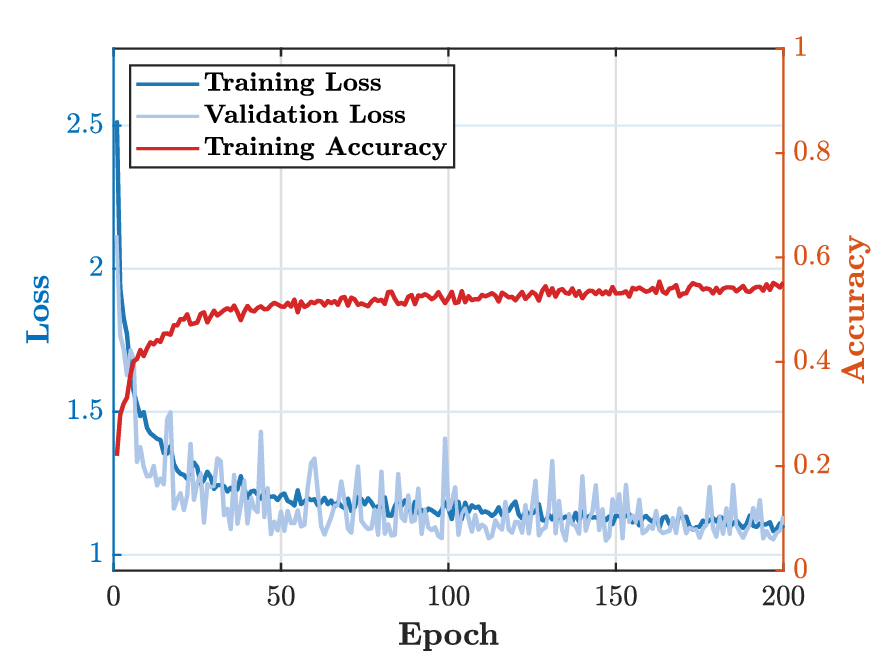}
        \caption{The curves of the training loss, validation loss, and training accuracy for BeamLLM.}
        \label{fig:converge}
    \end{figure}

    Fig. \ref{fig:converge} presents the training dynamics of BeamLLM over $200$ epochs, where the left vertical axis denotes the training and validation loss, and the right vertical axis represents the training accuracy. Note that training accuracy refers to the average top-$1$ accuracy in the training set. The convergence of both training and validation losses around similar levels suggests that the model has not severely overfitted. The gap between training and validation loss is not large, indicating that the model maintains a certain level of robustness on unseen data.

    \subsection{Standard Prediction}

    \begin{figure}[t]
        \centering
        \includegraphics[width=\linewidth]{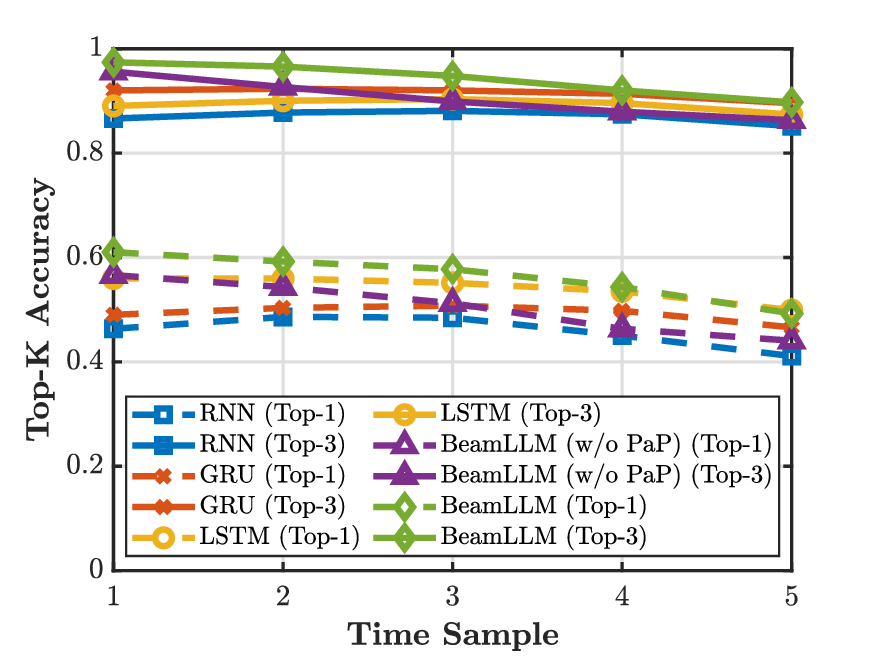}
        \caption{Top-$K$ accuracy performance of the proposed method comparing to several baselines in the standard prediction task.}
        \label{fig:n}
    \end{figure}

    \begin{figure}[t]
        \centering
        \includegraphics[width=\linewidth]{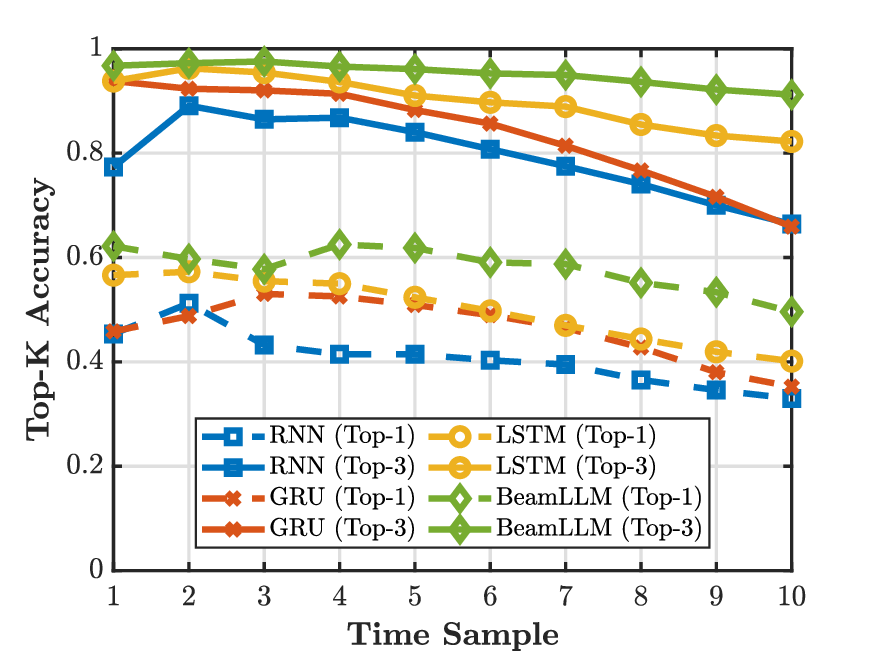}
        \caption{Top-$K$ accuracy performance of the proposed method comparing to several baselines in the few-shot prediction task.}
        \label{fig:f}
    \end{figure}
    In Fig. \ref{fig:n}, we present a comparative analysis of the top-$1$ and top-$3$ accuracy in the standard predictions across all models. Increasing $K$ improves top-$K$ accuracy, while as prediction horizon extends further into the future, the accuracy gradually decreases. Among the models, BeamLLM achieves the highest top-$1$ and top-$3$ accuracy scores, reaching $\mathbf{61.01\%}$ and $\mathbf{97.39\%}$, respectively. Additionally, as the number of time samples increases, the decay in the top-$K$ accuracy for the LSTM model is minimal. Specifically, the top-$1$ and top-$3$ accuracy only decrease by $\mathbf{6.03\%}$ and $\mathbf{1.65\%}$, respectively, across time samples ranging from $1$ to $5$. This smaller reduction highlights the adaptability of LSTMs, as their gating mechanism adjusts information retention and updating based on task demands.
    
    The results of the ablation study highlight the performance differences of the BeamLLM with and without the use of PaP. The average performance gap in top-$1$ accuracy between the two models is $\mathbf{5.81\%}$, while the gap in top-$3$ accuracy is $\mathbf{3.62\%}$. When comparing these scenarios, we observe that the integration of PaP significantly improves both performance and stability, compared to simply inputting the reprogrammed patch into the frozen LLM. This underscores the effectiveness of PaP in the context of this task.

    \subsection{Few-Shot Prediction}

    In Fig. \ref{fig:f}, we present the top-$1$ and top-$3$ accuracy performance for the few-shot forecasting task. Existing DL prediction methods perform poorly in this scenario, particularly as the prediction horizon extends, resulting in severe performance degradation. Even for the previously most stable LSTM model, during the progression from time sample $1$ to 10, the top-$1$ accuracy is decreased by $\mathbf{16.48\%}$, and the top-$3$ accuracy is decreased by $\mathbf{11.58\%}$. In contrast, BeamLLM significantly outperforms all baseline methods, with only $\mathbf{12.56\%}$ and $\mathbf{5.55\%}$ performance degradation, respectively. We attribute this superior performance to the successful activation of knowledge through the reprogrammed LLM.

    \subsection{Analysis of Complexity}
        All experiments are conducted in the same environment, specifically on Google Colab with an NVIDIA A$100$ GPU and $40$ GB of RAM. We investigate the training complexity and inference complexity of different models in terms of the number of trainable and non-trainable parameters, as well as the average inference time per sample (without YOLOv$4$ feature extraction module). Overall, the inference time of all models is significantly shorter than the sampling interval, which corresponds to $7.79$ frames per second (FPS) \cite{deepsense6g}. As shown in Table~\ref{tab:complexity}, although the backbone model in BeamLLM is frozen during training, the number of trainable parameters remains substantial. Additionally, its average inference time is higher than that of traditional models. While this introduces a higher deployment cost, it also suggests that the full potential of BeamLLM has yet to be fully realized and optimized.

\section{Conclusions}
    This work has presented an innovative BeamLLM for vision-empowered beam prediction, significantly improving accuracy and robustness in mmWave systems through reprogramming. Experimental results have highlighted LLMs’ superior contextual inference capabilities compared to conventional DL models in standard and few-shot prediction. 
    
    However, the performance gains come with increased inference complexity. The massive parameter scale of LLMs may introduce higher resource consumption and latency. Nevertheless, BeamLLM remains practical, particularly due to its exceptional few-shot prediction capability, which enables predictions over a longer horizon. Practical deployments require a trade-off between model complexity and real-time constraints, necessitating acceleration methods such as model compression or lightweight architecture design. By advancing these aspects, the proposed framework could serve as a scalable and efficient beam management solution for $6$G ISAC systems.

\balance

    \begin{table}[t]
        \caption{The number of model parameters and average inference time.}
        \centering
        \begin{tabular}{|c|c|c|c|}
        \hline 
        \textbf{Models} & \textbf{\makecell{\# total \\params.}} & \textbf{\makecell{\# trainable \\params.}} & \textbf{\makecell{Average inf. \\time (sec)}}\\
        \hline 
        RNN& $29,505$& $29,505$ &$8.0\times10^{-6}$\\
        \hline  
        GRU & $79,425$& $79,425$&$3.7\times 10^{-5}$\\
        \hline 
        LSTM& $104,385$ & $104,385$& $1.9\times10^{-5}$\\
        \hline
        BeamLLM & $178,303,798$& $53,863,990$  &  $7.5\times 10^{-4}$\\
        \hline 
        \end{tabular}
        \label{tab:complexity}
    \end{table}

\bibliographystyle{IEEEtran}
\bibliography{IEEEabrv, ref}
\end{document}